\documentclass[sigconf]{acmart}
\AtBeginDocument{%
  \providecommand\BibTeX{{%
    \normalfont B\kern-0.5em{\scshape i\kern-0.25em b}\kern-0.8em\TeX}}}

\usepackage{multirow}
\usepackage{multicol}
\usepackage{booktabs}
\usepackage{subfigure}
\usepackage{subcaption}
\usepackage{hhline}
\usepackage{color}
\usepackage{wasysym}
\usepackage{pifont}
\usepackage[ruled, vlined]{algorithm2e}
\usepackage{algpseudocode}
\usepackage{colortbl}
\usepackage{enumitem}
\usepackage{adjustbox}
\usepackage{graphicx}
\usepackage{caption} 

\newcommand{\figref}[1]{Figure~\ref{#1}}

\newcommand{\best}{\cellcolor[HTML]{DED0B6}}
\newcommand{\second}{\cellcolor[HTML]{FAEED1}}

\newcommand\model{\textsc{ZeroG}\xspace}
\newcommand\modelv{\textsc{ZeroG}}
\definecolor{high}{gray}{0.9}
\definecolor{LightCyan}{rgb}{0.88,1,1}
\definecolor{DarkBlue}{rgb}{0,0, 0.4}
\definecolor{color1}{RGB}{180, 96, 96} 
\definecolor{color2}{RGB}{33, 70, 199} 

\copyrightyear{2024}
\acmYear{2024}
\setcopyright{acmlicensed}\acmConference[KDD '24]{Proceedings of the 30th ACM SIGKDD Conference on Knowledge Discovery and Data Mining}{August 25--29, 2024}{Barcelona, Spain}
\acmBooktitle{Proceedings of the 30th ACM SIGKDD Conference on Knowledge Discovery and Data Mining (KDD '24), August 25--29, 2024, Barcelona, Spain}
\acmDOI{10.1145/3637528.3671982}
\acmISBN{979-8-4007-0490-1/24/08}

\begin{document}
\title{\textsc{ZeroG}: Investigating Cross-dataset Zero-shot Transferability in Graphs}

\makeatletter
\def\authornotetext#1{
	\g@addto@macro\@authornotes{%
	\stepcounter{footnote}\footnotetext{#1}}%
}
\makeatother

\newcommand{\paratitle}[1]{\vspace{0.8ex}\noindent \textbf{#1}}

\setcopyright{acmcopyright}
\copyrightyear{2024}
\acmYear{2024}

\acmConference[KDD'24]{KDD'24: SIGKDD Conference on Knowledge Discovery and Data Mining}{August 25-29, 2024}{Barcelona, Spain}
\acmBooktitle{KDD'24: SIGKDD Conference on Knowledge Discovery and Data Mining, August 25-29, 2024, Barcelona, Spain}

\authornotetext{Corresponding author.}

\author{Yuhan Li}
\affiliation{%
  \institution{HKUST (GZ)}
  \country{Guangzhou, China}
  }
\email{yli258@connect.hkust-gz.edu.cn}

\author{Peisong Wang}
\affiliation{
  \institution{THU}
  \country{Shenzhen, China}
  }
\email{wps22@mails.tsinghua.edu.cn}

\author{Zhixun Li}
\affiliation{
  \institution{CUHK}
  \country{Hong Kong SAR, China}
  }
\email{zxli@se.cuhk.edu.hk}

\author{Jeffrey Xu Yu}
\affiliation{%
  \institution{CUHK}
  \country{Hong Kong SAR, China}
  }
\email{yu@se.cuhk.edu.hk}

\author{Jia Li$^{*}$}
\affiliation{%
  \institution{HKUST (GZ)}
  \country{Guangzhou, China}
  }
\email{jialee@ust.hk}



\begin{abstract}
With the development of foundation models such as large language models, zero-shot transfer learning has become increasingly significant. This is highlighted by the generative capabilities of NLP models like GPT-4, and the retrieval-based approaches of CV models like CLIP, both of which effectively bridge the gap between seen and unseen data. In the realm of graph learning, the continuous emergence of new graphs and the challenges of human labeling also amplify the necessity for zero-shot transfer learning, driving the exploration of approaches that can generalize across diverse graph data without necessitating dataset-specific and label-specific fine-tuning. In this study, we extend such paradigms to  \underline{Zero}-shot transferability in \underline{G}raphs by introducing \underline{\model}, a new framework tailored to enable cross-dataset generalization. Addressing the inherent challenges such as feature misalignment, mismatched label spaces, and negative transfer, we leverage a language model to encode both node attributes and class semantics, ensuring consistent feature dimensions across datasets. We also propose a prompt-based subgraph sampling module that enriches the semantic information and structure information of extracted subgraphs using prompting nodes and neighborhood aggregation, respectively. We further adopt a lightweight fine-tuning strategy that reduces the risk of overfitting and maintains the zero-shot learning efficacy of the language model. The results underscore the effectiveness of our model in achieving significant cross-dataset zero-shot transferability, opening pathways for the development of graph foundation models\footnote{Codes and data are available at \url{https://github.com/NineAbyss/ZeroG}}.

\end{abstract}





\maketitle
\section{Introduction}
\label{sec:intro}

When encountering a completely new problem, humans typically start to compare and connect it with the knowledge that they are familiar with \cite{chen2023understanding}. The same idea is also applied to Machine Learning (ML). ML methods usually focus on classifying instances whose classes have already been seen in training. However, in practice, many applications necessitate classifying instances whose classes have not been seen previously, requiring models to utilize the knowledge gained before to begin reasoning and problem-solving. For example, a model trained on animal images may have never seen a ``zebra'' during its training. Yet, by understanding the concept of ``stripes'' and ``horse-like'' animals from its training, it can successfully identify a zebra when it encounters one for the first time. 

This trend of zero-shot capabilities in ML, particularly after the advent of foundation models such as large language models (LLMs), has demonstrated considerable advancements in the field of AI. Zero-shot learning is a powerful learning paradigm, in which the classes covered by training instances and the classes we aim to classify are disjoint. In the natural language processing (NLP) field, zero-shot learning is unified by a \textbf{\textit{generative}} paradigm, with LLMs such as GPT-4 \cite{achiam2023gpt} and LLaMA \cite{touvron2023llama} tackling new data they haven't been explicitly trained based on their extensive pre-training on massive corpora. While in the computer vision (CV) field, zero-shot learning relies on a \textbf{\textit{retrieval}} paradigm, where models like CLIP \cite{radford2021learning} map images to text in a shared space, allowing recognition of new images by their conceptual similarity to label descriptions.

Graphs are prevalent across multiple disciplines with diverse applications \cite{tang2022icmlGAD, tang2023gadbench,li2019semi,zheng2021convolutional,li2022devil,zhang2022knowledge,zhao2023effective,yang2022semantically}. However, we notice that graph learning faces two major challenges: \textbf{(1) \textit{the emergence of new graphs}}, which makes it impractical to train graph models like Graph Neural Networks (GNNs) \cite{kipf2016semi,velivckovic2017graph} on each individual graph; \textbf{(2) \textit{the difficulty of human labeling}} due to the complex and diverse nature of graph-structured data. Therefore, zero-shot learning is worth exploring in graph learning, as it allows graph models to generalize and perform reasoning over graphs that have never been seen before. This is also essential for reaching the goal of graph foundation models that can adapt to different data without extra fine-tuning. Due to the lack of detailed discussion and analysis on this task in prior work, our goal is to investigate zero-shot transferability in graphs for conducting cross-dataset transfers, which involves a pre-training phase with multiple datasets that have non-overlapping classes with the test datasets. According to recent surveys \cite{liu2023towards,li2023survey}, we consider this setting well-suited for the graph foundation model era, emphasizing broad generalization across different data sources. We offer a comprehensive discussion on how our setting differs with other related tasks in Section \ref{sec:pre}.

Graph transfer learning usually follows a ``pre-training and fine-tuning'' strategy, which aims to learn some general knowledge for the graph model with easily accessible information to reduce the annotation costs of new graphs \cite{zhao2024weakly}. Effective pre-training techniques include node-level comparison \cite{zhu2021graph} or reconstruction \cite{cheng2023oursgnn3,hou2022graphmae}, edge-level pretext like edge prediction \cite{jin2020self}, and graph-level contrastive learning such as GraphCL \cite{you2020graph}. However, to perform zero-shot transfer across datasets, the increased number of pre-training datasets and the absence of labels for the target dataset present unique challenges for these transfer learning methods.

\paratitle{Primary challenges.} Traditional GNNs face challenges in zero-shot transfer primarily due to dimension misalignment, mismatched label spaces, and negative transfer. Dimension misalignment arises when different datasets use varying shallow embedding techniques, such as bag-of-words or TF-IDF \cite{salton1988term}, leading to feature dimensions that are incompatible across datasets. This makes it difficult for a model trained on one dataset to adapt to another with different feature dimensions. Another issue is mismatched label spaces. A GNN trained on a dataset with a certain number of classes might fail on a dataset with a varying number of classes or where classes with the same count signify different meanings. In addition, negative transfer can also occur when GNNs overfit to the peculiarities of the training data, reducing its ability to generalize to new datasets with distinct structures or semantics \cite{sun2023all}. Recently, efforts have been made to use LLMs directly as classifiers for zero-shot inference in graphs \cite{chen2023exploring,huang2023can}. For example, for each node (\emph{i.e.,} one paper) in a citation network, we can directly input the title and abstract of the paper into a LLM and then ask which category the paper belongs to. However, it faces two main challenges. The first issue is data leakage. LLMs pre-trained on comprehensive text corpora may have been exposed to parts of the target datasets, possibly leading to an unfair advantage in their performance \cite{he2023explanations}. Secondly, it is hard for LLMs to incorporate graph structure, which is vital for graph tasks such as node classification that depend on both the attributes of the nodes and the connections between them \cite{li2023gslb}. We provide a more detailed discussion of the challenges in Section \ref{sec:challenges}.

\paratitle{Presented work.} We introduce \model, a framework designed for tackling cross-dataset zero-shot transfer in graphs. Specifically, to address the dimension misalignment challenge, we propose unifying the graph representation learning through a pre-trained language model (LM) to encode both text attributes associated with nodes and descriptions associated with classes. This approach maps the node and class features of various datasets to a unified semantic space of the same dimension, thereby addressing the issue of dimension misalignment in cross-dataset transfer. To address the issue of mismatched label spaces, we reformulate the node classification task into a text similarity task in both pre-training and inference phases. To tackle the negative transfer challenge, we introduce a prompt-based subgraph sampling strategy that enhances the general semantic information of datasets by prompting nodes while incorporating structural information through simple neighborhood aggregation. Additionally, we employ a lightweight fine-tuning approach to ensure suitability for pre-training datasets, minimizing the risk of overfitting and preserving the LM's zero-shot learning capabilities.

Our main contributions can be summarized as follows: (1) \textit{Comprehensive Analysis.} We are the first to systematically summarize and discuss the existing attempts and challenges associated with cross-dataset zero-shot transferability in graphs (Section \ref{sec:pre} and \ref{sec:challenges}). (2) \textit{Architecture Design.}  We proposed \modelv, a model designed for the lightweight training of language models to enable zero-shot transfer across diverse text-attribute graphs (Section \ref{sec:method}). (3) \textit{Superior Performance.} Our model has demonstrated effective zero-shot transferability across seven well-known benchmark datasets, \emph{i.e.,} it can even achieve results comparable to semi-supervised methods on Pubmed (Section \ref{sec:exp}).

\section{Preliminaries}
\label{sec:pre}

\paratitle{Notations.} To maintain consistency of notations, we use bold uppercase and lowercase letters to represent matrices and vectors, and calligraphic font types to denote sets. Given a Graph $\mathcal{G}=(\mathcal{V}, \mathbf{A}, \mathcal{T})$, where $\mathcal{V}=\{v_1, v_2, \ldots, v_N\}$ is the set of $N$ nodes; $\mathbf{A}\in\{0,1\}^{N\times N}$ is the adjacency matrix, if $v_i$ and $v_j$ are connected, $\mathbf{A}_{ij}=1$, otherwise $\mathbf{A}_{ij}=0$; $\mathcal{T}=\{t_1, t_2, \ldots, t_N\}$ is the set of node attributes, and each node $v_i$ is associated with attributes $t_i$. Typically, in most previous graph machine learning literature, such attribute information can be encoded into shallow embeddings $\mathbf{X}=[\mathbf{x}_1, \mathbf{x}_2, \ldots, \mathbf{x}_N]\in\mathbb{R}^{N\times f}$ by naive methods (\emph{e.g.}, bag-of-words or TF-IDF \cite{salton1988term}), where $f$ is the dimension of embeddings.

\paratitle{Existing Explorations.} Most of the existing works in node classification task use labeled nodes to train GNNs to predict the unlabeled nodes with the same label space of training set on a single graph, which we refer to \emph{in-dataset semi-supervised learning}. However, these works of literature ignore the generalization of GNNs to a completely new graph dataset. In recent years, several works have started to focus on the transferability of GNNs, and there are two popular scenarios: \emph{unsupervised graph domain adaptation (UGDA)} and \emph{in-dataset zero-shot}. Specifically, let $\mathcal{G}_s=(\mathcal{V}_s,\mathbf{A}_s,\mathbf{X}_s)$ and $\mathcal{G}_t=(\mathcal{V}_t, \mathbf{A}_t, \mathbf{X}_t)$ be source graph and target graph. \emph{UGDA} aims to learn a classification model on a fully labeled source graph $\mathcal{G}_s$, and accurately classify the nodes in the target graph $\mathcal{G}_t$ with the same label space, which can be denoted as $\mathcal{G}_s\cap\mathcal{G}_t=\emptyset, \mathcal{Y}_s=\mathcal{Y}_t$. While \emph{in-dataset zero-shot} focuses on transferability within a single graph, whose objective is categorizing unlabeled nodes into unseen classes within target label space $\mathcal{Y}_t$, given that all labeled nodes belong to the seen classes $\mathcal{Y}_s$, which can be denoted as $\mathcal{G}_s=\mathcal{G}_t, \mathcal{Y}_s\cap\mathcal{Y}_t=\emptyset$.

\begin{figure}[t]
    \centering
    \resizebox{\linewidth}{!}{
    \includegraphics{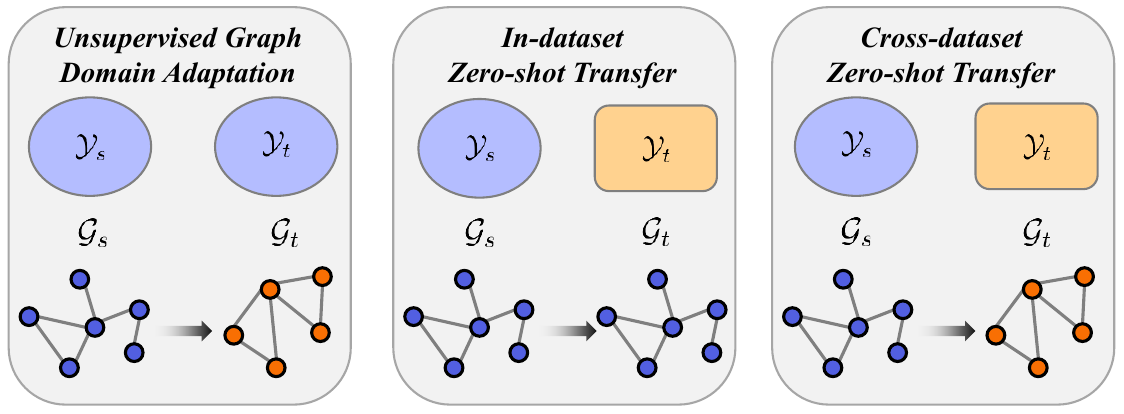}}
    \caption{Analysis of tasks associated with the zero-shot transfer in graphs.} 
    \label{fig:task}
\vspace{-4mm}
\end{figure}

\paratitle{Cross-dataset zero-shot.} Recently, a multitude of models that demonstrate cross-dataset transferability have emerged in fields like NLP and CV. In this work, we mainly focus on cross-dataset zero-shot node classification task. We aim to train a classification model learned on fully labeled source graph $\mathcal{G}_s$ and generate satisfactory predictions on a completely different target graph $\mathcal{G}_t$ with distinct label space $\mathcal{Y}_t$, which can be denoted as $\mathcal{G}_s\cap\mathcal{G}_t=\emptyset, \mathcal{Y}_s\cap\mathcal{Y}_t=\emptyset$. The difference between \emph{cross-dataset zero-shot} and the above two scenarios is illustrated in \figref{fig:task}. In fact, both \emph{UDGA} and \emph{in-dataset zero-shot} scenarios have significant limitations in practical applications. For instance, \emph{UGDA} requires consistent label space between the source and target graphs, making it unable to handle downstream inference with unseen classes. \emph{In-dataset zero-shot} can only perform zero-shot within a single graph, and such methods will become ineffective when encountering a completely new graph. \emph{Cross-dataset zero-shot} is a very practical scenario, but it is largely under-explored.

\section{Challenges}
\label{sec:challenges}

\emph{\ding{182} Why do GNNs fall short in cross-dataset zero-shot node classification?} 
\textbf{\textit{(1) Dimension misalignment}}. During pre-training on multiple datasets, aligning feature dimensions is crucial for graph models to handle data consistently. Currently, the use of shallow embeddings such as bag-of-words, skip-gram \cite{mikolov2013distributed}, and TF-IDF \cite{salton1988term} in mainstream benchmark datasets can lead to dimension misalignment issues across different source datasets (\emph{e.g.}, Cora with $1433$ dimensions, while Citeseer with $3703$ dimensions). This is hard to perform zero-shot transfer due to a graph model pre-trained on one dataset may struggle to process the other due to this inconsistency in feature dimension. In addition, dimension misalignment between source datasets and the target datasets can also make it challenging to apply the model directly to the downstream dataset, as the dimensions it has learned to process may not equal to those it encounters during transfer. \textbf{\textit{(2) Mismatched label spaces.}} A GNN's classification head, fixed to the number of classes seen during pre-training, cannot easily adapt to a different number of classes in the target dataset, leading to potential mismatched issues. Additionally, even when the number of labels is the same, labels may carry different meanings across datasets. For example, the categories in citation networks may not easily translate to those in web link datasets due to the distinct contexts they represent. \textit{\textbf{(3) Negative transfer.}} In graph data, negative transfer continues to be a topic of enduring discussion \cite{sun2023all,jin2020self,gao2024oursproteinprompt}. It often occurs when there is a significant disparity in structure or semantics between graphs. Fully adapting graph models to upstream datasets often causes overfitting, where the model becomes too specialized to the pre-training data's characteristics. This may hurt the model's performance on target datasets that differ in structure or meaning, as it may fail to capture the broader patterns necessary for zero-shot transfer learning.

\noindent
\emph{\ding{183} Why do large language models fall short in cross-dataset zero-shot node classification?} \textbf{\textit{(1) Data leakage.}} Data leakage in LLMs has become a focal point of discussion \cite{aiyappa2023can}. Given that LLMs such as GPT-4 \cite{achiam2023gpt} and LLaMA \cite{touvron2023llama} undergo pre-training on extensive text corpora, it's likely that they may have seen and memorized at least part of the test data of the common benchmark datasets. This can lead to an overestimation of the capabilities of language models during evaluation \cite{huang2023can,chen2023exploring,he2023explanations}. \textbf{\textit{(2) Lack of structural information.}} In the node classification task, the prediction of a node is determined by both its own features and the contextual subgraph. However, language models typically rely on a textual item for prediction and do not consider structural information, which will jeopardize the utility of language models in node classification task. In recent years, several works have designed an auxiliary task (\emph{e.g.}, neighbor prediction or link prediction) to provide structural awareness capability to language models \cite{chien2021node, duan2023simteg}. Although they can effectively improve the accuracy of node classification on text-attributed graphs, they all require employing a GNN on top of the language model for prediction, which limits their ability to perform cross-dataset zero-shot learning. In this paper, we attempt to tackle the aforementioned issues and strive to mitigate them.
\section{Methodology}
\label{sec:method}

\begin{figure*}[t]
    \centering
    \resizebox{\linewidth}{!}{
    \includegraphics{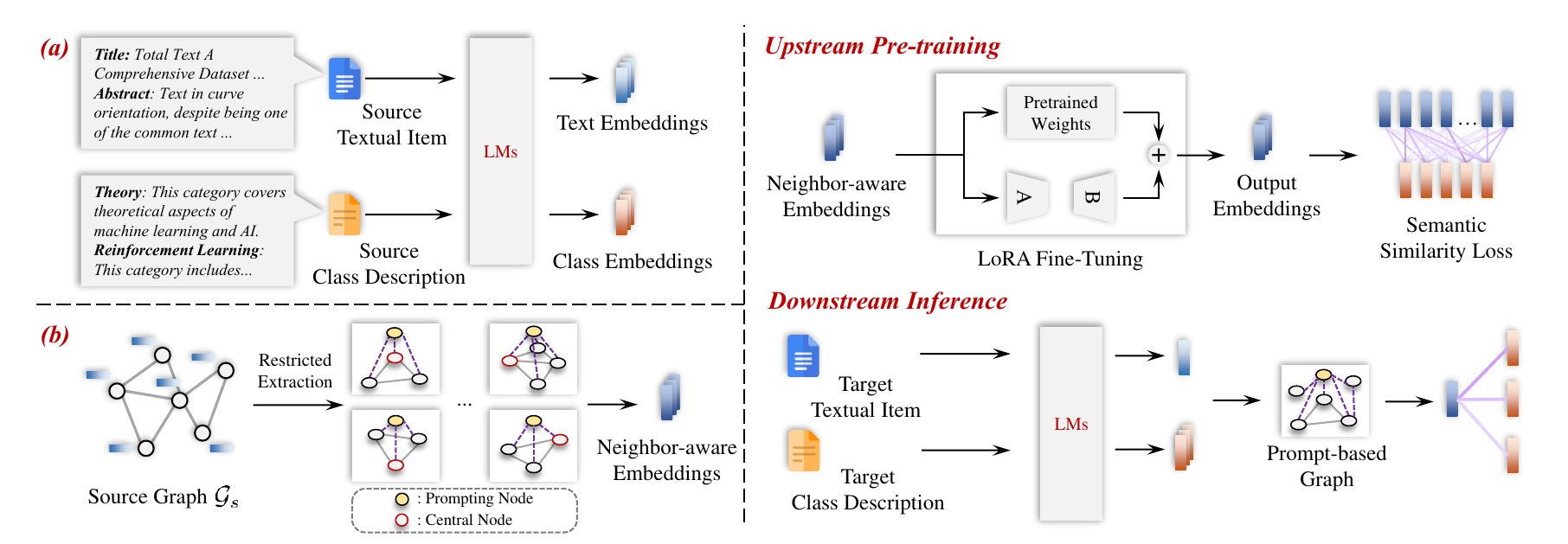}}
    \caption{Our proposed pipeline \model facilitates cross-dataset zero-shot node classification with three key components: (a) It uses \underline{unified graph representations} to merge node and class encodings via language models. (b) It employs \underline{prompt-based subgraph sampling} to create pre-training data from rich subgraphs. (c) Its \underline{upstream pre-training} adopts a parameter-efficient approach to suit various datasets while preserving zero-shot abilities and preventing overfitting. Finally, we can leverage the pre-trained model to perform \underline{downstream inference} on target datasets. } 
    \label{fig:pipeline}
    \vspace{-1mm}
\end{figure*}

Our proposed \model posits a paradigm shift in zero-shot graph learning, drawing inspiration from the adaptive nature of LLMs. By fine-tuning a relatively small language model in a low-resource setting on source datasets for pre-training, \model is capable of exhibiting substantial zero-shot learning abilities on downstream target datasets. Figure \ref{fig:pipeline} illustrates the pipeline of \modelv, which can be divided into three parts. Firstly, we integrate the encoding of node features and class descriptions within the graph via employing language models, achieving a standardized representation framework. In the second part, we extract diverse subgraphs from source datasets and organically integrate dataset-specific information into subgraphs leveraging a graph prompting paradigm. Finally, a lightweight alignment projector is adapted to map source graph features into the original word embedding space.

\subsection{Unified Graph Representation}
\label{subsec:unified}

The most mainstream benchmark datasets (\emph{e.g.}, Cora and Pubmed) often adopt naive methods to encode node attribute information using shallow embeddings, such as bag-of-words, skip-gram \cite{mikolov2013distributed}, or TF-IDF \cite{salton1988term}. However, in the zero-shot setting, traditional shallow embeddings often encounter issues with feature dimension alignment across different datasets (\emph{e.g.}, Cora has 1433 dimensions, while Citeseer has 3703 dimensions). This misalignment leads to two main challenges: (1) \emph{\textbf{Inconsistency in pre-training}}. During the pretraining phase, aligning different source datasets becomes problematic due to varying feature dimensions, hindering the development of a coherent model understanding. (2) \emph{\textbf{Difficulty in applying to target datasets}}. The lack of uniformity in feature dimensions between source and target datasets poses significant challenges in applying pre-trained models effectively to new, unseen data. 

To address the two challenges mentioned above, we employ a unified pre-trained LM to encode both node attributes and descriptions associated with classes. Specifically, let $t_n$ denote the node attributes of $v$, $t_c$ denote the descriptions of class $c$, and \textsc{LM} be a pre-trained LM. The representations of node attributes and class descriptions can be obtained by applying the \textsc{LM} as follows:
\begin{equation}
    \textbf{h}_n = \textsc{LM}(t_n) \in \mathbb{R}^d, \quad \textbf{h}_c = \textsc{LM}(t_c) \in \mathbb{R}^d,
\end{equation}

\noindent where $\textbf{h}_n$ and $\textbf{h}_c$ represent the outputs from the final hidden layer corresponding to the \textsc{[CLS]} token for inputs $t_n$ and $t_c$, respectively, and $d$ denotes the dimension of the hidden layer in the LM. An example of node features and class descriptions is shown in Appendix \ref{app:node_des}. By doing so, we map the node and class features of various datasets to a unified semantic space and the same dimension $d$, addressing the issue of dimension misalignment in cross-dataset transfer. 

\subsection{Prompt-based Subgraph Sampling}

To facilitate zero-shot transfer capabilities in target datasets, transferring both structural and semantic information from the source datasets is crucial, ensuring that target datasets are not only enriched with the foundational structural patterns but also imbued with the contextual semantic nuances from the source \cite{tang2023graphgpt}. In \modelv, we introduce a novel prompt-based subgraph sampling strategy, which captures essential structural and semantic features by extracting subgraphs from source graphs to construct the pre-training set and introduce a prompt node that enriches the unique semantics of these source datasets.

\subsubsection{Restricted Extraction}

Given that source datasets can be numerous and varied, with significant variations in graph size, density, node types, edge types, and overall topology, we employ restricted subgraph extraction to avoid overly simplistic subgraphs and limit the number of subgraphs extracted. For example, a subgraph that is too small will not provide enough structural information for dataset transfer. Similarly, if a subgraph only represents few classes, even a single class, it may fall short in offering the diverse semantic guidance needed for transfer. 

For each source dataset, we iteratively extract $k$-hop subgraphs using every node as a center. Adjusting $k$ provides flexibility in controlling both individual subgraph sizes and the overall number of subgraphs, tailored to the dataset's specific characteristics. Next, we apply a class-based filtering criterion in which we only consider those subgraphs where the number of distinct classes surpasses half of the total classes in the dataset, i.e., for a subgraph $s$ of source dataset $v$, it holds that $|C_s^v| \geq \frac{1}{2} |C^v|$, where $|C_s^v|$ denotes the cardinality of the set of distinct classes in $s$ and $|C^v|$ denotes the cardinality of the set of all classes in $v$. This criterion guarantees that each subgraph meaningfully reflects the dataset's class diversity, facilitating a thorough transfer of the source dataset's fundamental semantic content. 

\subsubsection{Prompting Node}
\label{subsubsec:prompting}

Graph prompt techniques have been extensively explored for low-resource graph learning. For instance, graph prompts leveraged in well-known graph transfer models such as ProG \cite{sun2023all} and GraphPrompt \cite{liu2023graphprompt} are randomly initialized and then made learnable, adapting specifically to the source dataset they are connected with.  However, existing graph prompting methods struggle in zero-shot scenarios because inserting a trained dataset-specific prompt graph into a target graph is impractical, especially when dealing with multiple source datasets. In addition, the initial random state of graph prompts lacks the rich semantic information that characterizes both the source datasets and target datasets. 

In \modelv, we introduce semantic-enhanced prompting nodes as unique identifiers that carry general knowledge relevant to specific datasets. To create the prompting node for each dataset, we design a general description that serves as the text attribute for the corresponding prompting node, encompassing the fundamental attributes of the dataset it represents. More details about the dataset descriptions can be found in Appendix \ref{app:dataset_des}. We then integrate the prompting node into each extracted subgraph in the pre-training set as a unique identifier, providing more general semantics of the source dataset. The insertion pattern we employ is fully connected, that is, for each subgraph, the prompting node of the source dataset to which the subgraph belongs is connected to all nodes within the subgraph. The representation of the prompting node is initialized by the same LM described in Section \ref{subsec:unified}.

\subsubsection{Neighborhood Aggregation}
\label{subsubsec:neighbor}

Formally, supposing that one subgraph has adjacency matrix $\textbf{A}$, degree matrix $\textbf{M}$ and feature matrix $\textbf{H}$ processed by LM. Neighborhood aggregation involves the normalization of the adjacency matrix and subsequent aggregation. Initially, a self-loop is added to each node in the graph. This step integrates the node's own features with those of its neighbors. Subsequently, the normalized adjacency matrix of the subgraph is then calculated as:
$\textbf{A}_{\text{norm}} = \textbf{M}^{-\frac{1}{2}} \textbf{A} \textbf{M}^{-\frac{1}{2}}$. Using this normalized adjacency matrix, neighborhood aggregation is performed iteratively. In each iteration, the node features are updated by aggregating features from their immediate neighbors, including the nodes themselves. This process is mathematically represented as
$\textbf{H}^{(t+1)} = \textbf{A}_{\text{norm}} \textbf{H}^{(t)}$, where $\textbf{H}^{(t)}$ denotes the feature matrix at iteration $t$. The iteration continues for a predefined number of steps, allowing the node embeddings to progressively integrate information from their extended neighborhood.

\subsection{Upstream Pre-training}

One significant concern of pre-training LMs for source dataset compatibility is the resource-intensive nature of conventional full-parameter pre-training \cite{chen2023deepzero}, which also carries a substantial risk of overfitting, especially in zero-shot scenarios \cite{chen2023large,chen2024entity}. To circumvent these limitations, we utilize LoRA \cite{hu2021lora}, a parameter-efficient pre-training strategy, which ensures suitability for upstream datasets while minimizing overfitting risks and preserving the LM's zero-shot learning abilities.

LoRA injects trainable low-rank matrices into transformer layers of LM to approximate the weight updates. Consider a pre-trained weight matrix $\textbf{W} \in \mathbb{R}^{d \times k}$. LoRA updates this matrix with a low-rank decomposition expressed as $\textbf{W} + \Delta W = \textbf{W} + \textbf{W}_{\text{down}} \textbf{W}_{\text{up}}$, where $\textbf{W}_{\text{down}} \in \mathbb{R}^{d\times r}$ and $ \textbf{W}_{\text{up}} \in \mathbb{R}^{r\times k}$ are the trainable low-rank matrices. Specifically, we apply these updates to the query and value projection matrices, $\textbf{W}_q$ and $\textbf{W}_v$, within the multi-head attention sub-layer. For an input $\textbf{x}$ to this linear projection, LoRA modifies the resulting projection output $\textbf{h}$ in the following:
\begin{equation}
    \textbf{h} = \textbf{h} + \alpha \cdot\textbf{x}\textbf{W}_{\text{down}}\textbf{W}_{\text{up}},
\end{equation}
\noindent where $\alpha  \ge 1$ is a tunable scalar hyperparameter. Based on LoRA, we utilize a cross-entropy loss for pre-training. Formally, denoting the total pre-training set as $\mathcal{T}_{\text {pre }}$ and the node set of the subgraph $s$ as $N_s$, the loss function is defined as follows:
\begin{equation}
    \mathcal{L}_{\text {pre }}(\Theta) = - \sum_{s \in \mathcal{T}_{\text {pre }}} \sum_{n \in N_s} \log \frac{\exp \left(\operatorname{sim}\left(\textbf{h}_n, \textbf{h}_{y_n}\right)  \right)}{\sum_{c \in Y_s} \exp \left(\operatorname{sim}\left(\textbf{h}_n, \textbf{h}_c\right)  \right)},
\end{equation}

\noindent where $y_n$ represents the actual label of node $n$, and $Y_s$ represents the set of classes associated with subgraph $s$.  $\operatorname{sim}(\cdot)$ is a similarity measurement used to calculate the semantic similarity between node embeddings and class embeddings, where we employ the dot product here. Note that the loss is parameterized by $\Theta$, representing the weights of the low-rank matrices. The goal of learning is to minimize the loss $\mathcal{L}_{\text {pre}}$.

    

\subsection{Downstream Inference}

After pre-training on source datasets, we acquire a pre-trained LM enriched with upstream semantics and structure. This LM is designed for zero-shot transferability, allowing direct application to target datasets without any fine-tuning. We try to unify the pre-training and inference stages, enabling effective knowledge transfer as the datasets in the two phases are made more compatible by following a common template. However, it is still important to distinguish different downstream datasets, in order to capture dataset individuality and achieve dataset-specific optimum. 


Briefly, given a target dataset, we first apply the LM to generate embeddings for its nodes and classes, respectively. Creating a universal prompting node and updating node embeddings through neighborhood aggregation allows our LM to capture the target dataset's unique global context and local structures. Finally, the class that yields the highest similarity score is predicted to be the class of the node, which can be formalized as follows:
\begin{equation}
\label{eq:argmax}
y' = \text{argmax}_i (\text{sim}(\textbf{h}_n, \textbf{h}_{c_i}) \ | \ i \in \{1, \ldots, N\}),
\end{equation}

\noindent where $y'$ is the predicted label of the node $n$ and $N$ denotes the total number of classes. For $\operatorname{sim}(\cdot)$, we also use the dot product as the similarity function here.
\section{Experiments}
\label{sec:exp}

We evaluate \textsc{ZeroG} on real-world datasets for node classification to assess its performance in enhancing cross-dataset zero-shot graph transfer learning. In particular, we wish to answer the following research questions: \textbf{Q1}: How effective is \textsc{ZeroG} in a zero-shot learning scenario when the source and target datasets are within the same domain? \textbf{Q2}: How adaptable is our model when applied across different domains? \textbf{Q3}: How do the main components of our model impact the performance? \textbf{Q4}: What is the impact of hyper-parameter on performance? \textbf{Q5}: How does \textsc{ZeroG} compare to state-of-the-art methods in training efficiency? 

\subsection{Experimental Protocols}
\subsubsection{Datasets}
Our experiments are conducted on several public graph datasets. First we consider four commonly used \textit{citation networks}, \emph{i.e.,} Cora \cite{mccallum2000automating}, Citeseer \cite{giles1998citeseer}, Pubmed \cite{sen2008collective}, and ogbn-arxiv \cite{hu2020open}. To ensure data distribution practices in NLP and CV, where source datasets are typically larger than target datasets, we do not consider ogbn-arxiv as a target dataset but only employ it as one of the source datasets. Apart from this, to further explore in-domain transfer, we develop two \textit{co-purchase networks} from ogbn-products \cite{hu2020open}, named P-Home and P-Tech. We carefully curate these datasets to ensure that they are non-overlapping with respect to their classes. More details about these two datasets can be found in Appendix \ref{app:datasets}. In addition, in cross-domain transfer, we introduce another \textit{web link} dataset Wiki-CS \cite{mernyei2020wiki} to provide additional domain variance.  For a fair comparison, we follow OFA \cite{liu2023one} to provide Cora, Pubmed, ogbn-arxiv, and Wiki-CS with texts, both for nodes and classes. For CiteSeer, P-Home, and P-Tech, we use raw texts processed by Chen et al. \cite{chen2023exploring}. 

\begin{table}[t]
\centering
\caption{Statistics of datasets. }
\label{tab:data}
\resizebox{0.485\textwidth}{!}{
\begin{tabular}{@{}l|c|c|c|c|c@{}}
\toprule
Dataset  & \# Domain & \# Nodes & \# Edges  & \# Avg. D& \# Classes           \\ \midrule \midrule
Cora    & Citation & 2,708    & 5,429     & 4.00 & 7    \\
Citeseer & Citation & 3,186   & 4,277      &2.57  & 6  \\
Pubmed  & Citation & 19,717   & 44,338    &  4.50    & 3    \\ 
ogbn-arxiv    & Citation &  169,343   &   1,166,243   &  13.77  & 40    \\
P-Home   & Co-purchase & 9,790 &131,841  &26.93  &5  \\
P-Tech   & Co-purchase & 47,428& 2,077,241 & 87.60 &3    \\
Wiki-CS   & Web link & 11,701 & 216,123 &36.94 & 10   \\

\bottomrule
\end{tabular}
}
\vspace{-1mm}
\end{table}

\subsubsection{Baselines}

We note that few methods can be directly employed to perform zero-shot transfer across different graphs. In this case, we make adaptations of commonly used models or techniques with zero-shot capabilities, to serve as our baselines. Our baselines can be organized into three main categories: (1) \textbf{Graph self-supervised learning-based methods.} Graph SSL focuses on the transferability of learned \textit{structures} without the need of labels. We here consider three prominent graph SSL approaches including DGI \cite{velivckovic2018deep}, GraphCL \cite{you2020graph}, and GraphMAE \cite{hou2022graphmae}. For each target dataset, we employ GNN trained with SSL techniques on its own structure and then apply Eq. \ref{eq:argmax} to perform zero-shot classification, without an explicit classification head. (2) \textbf{Semantic similarity-based methods.} Intuitively, we can perform zero-shot classification by directly calculating the \textit{semantic} similarity between each node and class and then choosing the class with the highest similarity as the prediction for that node. We leverage four widely used LM as the baseline encoders, such as BERT \cite{devlin2018bert}, RoBERTa \cite{liu2019roberta}, E5 \cite{wang2022text}, and SentenceBert \cite{reimers2019sentence}. (3) \textbf{Graph foundation-based methods.} OFA \cite{liu2023one} proposes a unified way for cross-dataset transfer on graph data. It represents all nodes and edges as human-readable texts and utilizes a single LM to embed these textual descriptions from various datasets into a shared embedding space. A single GNN is then pre-trained on multiple graphs, and its capability is evaluated in zero-shot learning scenarios. Additionally, we report semi-supervised results on our dataset with GCN \cite{kipf2016semi} and GAT \cite{velivckovic2017graph}.

\subsubsection{Implementations}

To make a fair comparison, we follow OFA \cite{liu2023one} to utilize SentenceBert \cite{reimers2019sentence} as our LM. The dimension $d$ of both node and class representations is 768, and Adam \cite{kingma2014adam} with an initial learning rate 1e-4 is utilized as the optimizer. We apply a weight decay of 0.01 during pre-training. The maximum sequence length is set to 256. The rank and scaling factor of the LoRA adapter are set to 4 and 16, respectively. Dropout with a probability of 0.1 is used to alleviate over-fitting. The batch size is set to 1 since each batch processes a single subgraph. The iterations of neighborhood aggregation $\lambda$ and the hops of extracted subgraph $k$ are variable, depending on the size and sparsity of the dataset. It is noted that since we do not have a validation set, we fix the number of epochs (\emph{i.e.,} 3) for all datasets to ensure a fair comparison. All experiments are implemented by PyTorch Framework with a single NVIDIA A800 (80G) GPU. 

\begin{table}[t]
\caption{Test accuracy (\%) on target graphs for in-domain transferability. For each citation network (resp. co-purchase network) as target dataset, we consider the other citation networks (resp. co-purchase networks) as source datasets. $\mathcal{A}$ and $\mathcal{S}$ indicate whether structural and semantic information are used, respectively. GCN* and GAT* report results under the semi-supervised setting, which, except for P-Home and P-Tech, are all directly obtained from the original papers.}
\resizebox{1\linewidth}{!}{
\begin{tabular}{l|cc|ccc|cc}
\toprule
Methods      & $\mathcal{A}$ & $\mathcal{S}$ & Cora & Pubmed & Citeseer & P-Home & P-Tech \\ \midrule \midrule
\multicolumn{8}{c}{\textit{zero-shot settings}} \\ \midrule 
DGI \cite{velivckovic2018deep}         &     \ding{51}     &  \ding{55}        &  19.97    &  43.89        &   21.12     & 33.06      & 55.83 \\
GraphCL \cite{you2020graph}     &    \ding{51}       &    \ding{55}      &    26.22  &  43.73        &    20.59    & 37.44    &62.63  \\
GraphMAE \cite{hou2022graphmae}    &     \ding{51}      &  \ding{55}        &   34.79   &  \second{48.23}        &   34.62     & 37.04       &\second{73.37} \\ \midrule
BERT \cite{devlin2018bert}   &    \ding{55}       &   \ding{51}     &  19.90 & 34.79	& 23.76	  &   37.32   &56.44    \\
RoBERTa \cite{liu2019roberta}         &     \ding{55}      &     \ding{51}     &    28.91  &     27.33     &    30.95    & 35.50   &  66.31  \\ 
E5 \cite{wang2022text}           &      \ding{55}     &   \ding{51}       & 39.70 & 41.93	 & 45.89	 &57.56&59.17 \\ 
Sent-BERT \cite{reimers2019sentence} &    \ding{55}       &    \ding{51}     &  \second{52.25} &  41.71 
 & \second{47.52}  &  \second{63.22}  &67.21 \\
\midrule
OFA     \cite{liu2023one}      &   \ding{51}        &     \ding{51}    & 27.07  & 37.87   & 37.92   &32.86   & 71.03 \\ 
\midrule
\textbf{\model} (ours)        &   \ding{51}        &  \ding{51}       &   \best{\textbf{68.72}}    &   \best{\textbf{78.02}}       &   \best{\textbf{64.94}}     &\best{\textbf{73.20}}     & \best{\textbf{82.96}} \\ \midrule \midrule
\multicolumn{8}{c}{\textit{semi-supervised settings}} \\ \midrule
GCN*  \cite{kipf2016semi}         &     -      &    -      &  81.50    &   79.00       &   70.30     & 73.85     &93.28 \\ 
GAT*    \cite{velivckovic2017graph}    &  -         &     -     &   83.00   &     79.00     &   72.50     &73.46      &88.89   \\ 
\bottomrule
\end{tabular}}
\label{tab:main}
\vspace{-2mm}
\end{table}

\subsection{In-domain Transferability (RQ1)}

We first consider transferring within citation networks or co-purchase networks. The results are reported in Table \ref{tab:main} where for each citation network (resp. co-purchase network) as the target dataset, we use the other citation networks (resp. co-purchase networks) as the source datasets. It is observed that \model achieves significant performance gains on all the target datasets within both citation networks and co-purchase networks compared to baseline models, demonstrating the efficacy of our model. Remarkably, \model demonstrates its effectiveness by achieving 78.02\% accuracy on the Pubmed dataset. This performance is even on par with that of two semi-supervised learning methods, both of which boast an accuracy of 79.00\%.  Compared to methods that focus solely on structure, such as graph SSL-based methods, and those that rely purely on text similarity, like semantic similarity-based methods, our model achieves a comprehensive utilization of both structural and semantic information to facilitate zero-shot transfer. OFA \cite{liu2023one}, however, does not perform as well as our model or even other baselines that concentrate on either structures or semantics alone. The possible reasons are two-fold. Firstly, by freezing the LM parameters, OFA restricts the semantic space and domain-specific tuning, thereby failing to achieve satisfactory zero-shot transferability. Secondly, OFA employs a generic linear layer for all classes, that is, the probability associated with each node for every class is derived through shared parameters. Since class semantics are often unique, this shared approach can result in insufficient variation between classes. In contrast, \model computes dot product similarity between nodes and classes directly, preserving the inherent characteristics of each class. We also note that a higher number of pre-training classes relative to target classes, such as transferring from Arxiv+Cora+Citeseer to Pubmed or from P-Home to P-Tech, typically leads to notably better transfer performance, indicating that source data diversity enhances zero-shot transfer.

\begin{table}[]
\caption{Test accuracy (\%) on target graphs for cross-domain transferability. We use red and blue to signify performance increases or decreases, respectively, when pre-training with graphs from other domains versus in-domain (In-D) transfer. }
\label{tab:cross-domain}
\begin{adjustbox}{width=0.48\textwidth}
\begin{tabular}{l|l|ccc}
\toprule
Test      & Pre-training  & \multicolumn{1}{l}{OFA} & In-D & \modelv \\ \midrule \midrule
\multicolumn{1}{l|}{\multirow{1}{*}{Wiki-CS}}     & Arxiv $\cup$ Cora $\cup$ Pubmed $\cup$ Citeseer & 48.42 & - & 53.28  \\ \midrule

\multicolumn{1}{l|}{\multirow{1}{*}{Wiki-CS}}     & P-Home $\cup$ P-Tech &21.09  & - & 60.97  \\ \midrule

\multicolumn{1}{l|}{\multirow{1}{*}{Cora}}     & P-Home $\cup$ P-Tech  & 18.57 & 68.72 & 67.65\textcolor{color2}{\scriptsize{(-1.07\%)}} \\ \midrule

\multicolumn{1}{l|}{\multirow{1}{*}{Pubmed}}     & P-Home $\cup$ P-Tech  & 31.89 & 78.02 & 69.12\textcolor{color2}{\scriptsize{(-8.90\%)}} \\ \midrule

\multicolumn{1}{l|}{\multirow{1}{*}{Citeseer}}     & P-Home $\cup$ P-Tech  &20.78  & 64.94 & 53.17\textcolor{color2}{\scriptsize{(-11.77\%)}} \\ \midrule

\multicolumn{1}{l|}{\multirow{1}{*}{P-Home}}     & Arxiv $\cup$ Cora $\cup$ Pubmed $\cup$ Citeseer & 35.73 & 73.20 & 71.45\textcolor{color2}{\scriptsize{(-1.75\%)}} \\ \midrule

\multicolumn{1}{l|}{\multirow{1}{*}{P-Tech}}     & Arxiv $\cup$ Cora $\cup$ Pubmed $\cup$ Citeseer &  62.10 & 82.96 & 83.20\textcolor{color1}{\scriptsize{(+0.24\%)}}\\ 

\bottomrule
\end{tabular}
\end{adjustbox}
\end{table}

\subsection{Cross-domain Transferability (RQ2)}

We then explore the more challenging cross-domain transfer setting, where the difficulty arises from the larger variance in the underlying data-generating distributions and semantics between domains compared to in-domain transfer. The results are presented in Table \ref{tab:cross-domain}. \model consistently outperforms the baseline model OFA \cite{liu2023one} across all datasets, demonstrating robust cross-dataset transfer capabilities. Specifically, for the web link dataset Wiki-CS, which is utilized for testing cross-domain transfer, we notice that pre-training on co-purchase networks yields better results than pre-training on citation networks. This indicates intrinsic properties of co-purchase networks may be more transferable or relevant to Wiki-CS. Additionally, compared with in-domain transfer results, it is reasonable that there is a performance drop for almost all datasets due to the gaps between domains. Some datasets such as Cora and P-Home are less impacted by domain transfer, with decreases of 1.07\% and 1.75\% respectively compared to in-domain transfer. On the other hand, datasets like Citeseer and Pubmed are significantly affected, showing decreases of 11.77\% and 8.90\%, respectively. This reveals that they rely more heavily on domain-specific structures and knowledge for reasoning.

\subsection{Ablation Studies (RQ3)}
\label{sec:ablation}
We conduct ablation studies of our proposed \model to validate its key components, and present the results in Table \ref{tab:ablation}. The variant ``w/o $p$'' refers to our method without prompting node used during subgraph extraction introduced in Section \ref{subsubsec:prompting}; ``w/o NA'' denotes our method without neighborhood aggregation discussed in \ref{subsubsec:neighbor}; ``w/o norm'' indicates the absence of any normalizations within the subgraph; ``w/o LoRA'' means updating all parameters of the LM. From Table \ref{tab:ablation} we can see that all these components significantly contribute to the final results. Specifically, the integration of the prompting node not only serves as a unique identifier for the subgraph but also supplements it with general semantic context, leading to performance improvements across all datasets. The absence of neighborhood aggregation decreases the performance, for example, a 25.43\% decrease in Cora and a 30.81\% decrease in Pubmed, which indicates the importance of leveraging structural information across datasets. Interestingly, normalization appears to be a critical factor, which may be attributed to its role in adjusting for different subgraph sizes and densities, enabling the model to learn in a generalizable manner. Moreover, ``w/o LoRA'' shows that allowing full parameter updates can lead to severe overfitting. For example, on the Cora and Citeseer datasets, we observe performance declines of 51.36\% and 40.96\%, respectively, demonstrating the effectiveness of LoRA in such scenarios.

\begin{table}[t]
\caption{Ablation study for \textsc{ZeroG}.}
\label{tab:ablation}
\begin{adjustbox}{width=0.48\textwidth}
\begin{tabular}{l|ccc|cc}
\toprule
Methods     & Cora  & Pubmed & Citeseer & P-Home & P-Tech \\ \midrule \midrule

\textsc{ZeroG} & 68.72 & 78.02 & 64.94 & 73.20 & 82.96 \\ \midrule

- (w/o $p$) & 68.25\textcolor{color2}{\scriptsize{(-0.47\%)}} &  76.49\textcolor{color2}{\scriptsize{(-1.53\%)}}  & 61.64\textcolor{color2}{\scriptsize{(-3.30\%)}} & 70.46\textcolor{color2}{\scriptsize{(-2.74\%)}} & 79.68\textcolor{color2}{\scriptsize{(-7.18\%)}} \\ \midrule

- (w/o NA)  & 43.31\textcolor{color2}{\scriptsize{(-25.43\%)}} & 47.21\textcolor{color2}{\scriptsize{(-30.81\%)}} & 48.68\textcolor{color2}{\scriptsize{(-16.26\%)}} & 60.26\textcolor{color2}{\scriptsize{(-12.94\%)}} & 58.91\textcolor{color2}{\scriptsize{(-27.95\%)}} \\ \midrule

- (w/o norm) & 54.43\textcolor{color2}{\scriptsize{(-13.39\%)}} & 39.25\textcolor{color2}{\scriptsize{(-38.77\%)}} & 34.84\textcolor{color2}{\scriptsize{(-30.10\%)}} & 41.26\textcolor{color2}{\scriptsize{(-31.94\%)}} & 72.17\textcolor{color2}{\scriptsize{(-14.69\%)}} \\ \midrule

- (w/o LoRA)  & 17.36\textcolor{color2}{\scriptsize{(-51.36\%)}} & 46.49\textcolor{color2}{\scriptsize{(-31.35\%)}} & 23.98\textcolor{color2}{\scriptsize{(-40.96\%)}} & 39.77\textcolor{color2}{\scriptsize{(-33.43\%)}} & 87.22\textcolor{color1}{\scriptsize{(+4.26\%)}} \\

\bottomrule
\end{tabular}
\end{adjustbox}
\end{table}

\begin{figure}[t]
    \centering
    \resizebox{\linewidth}{!}{
    \includegraphics{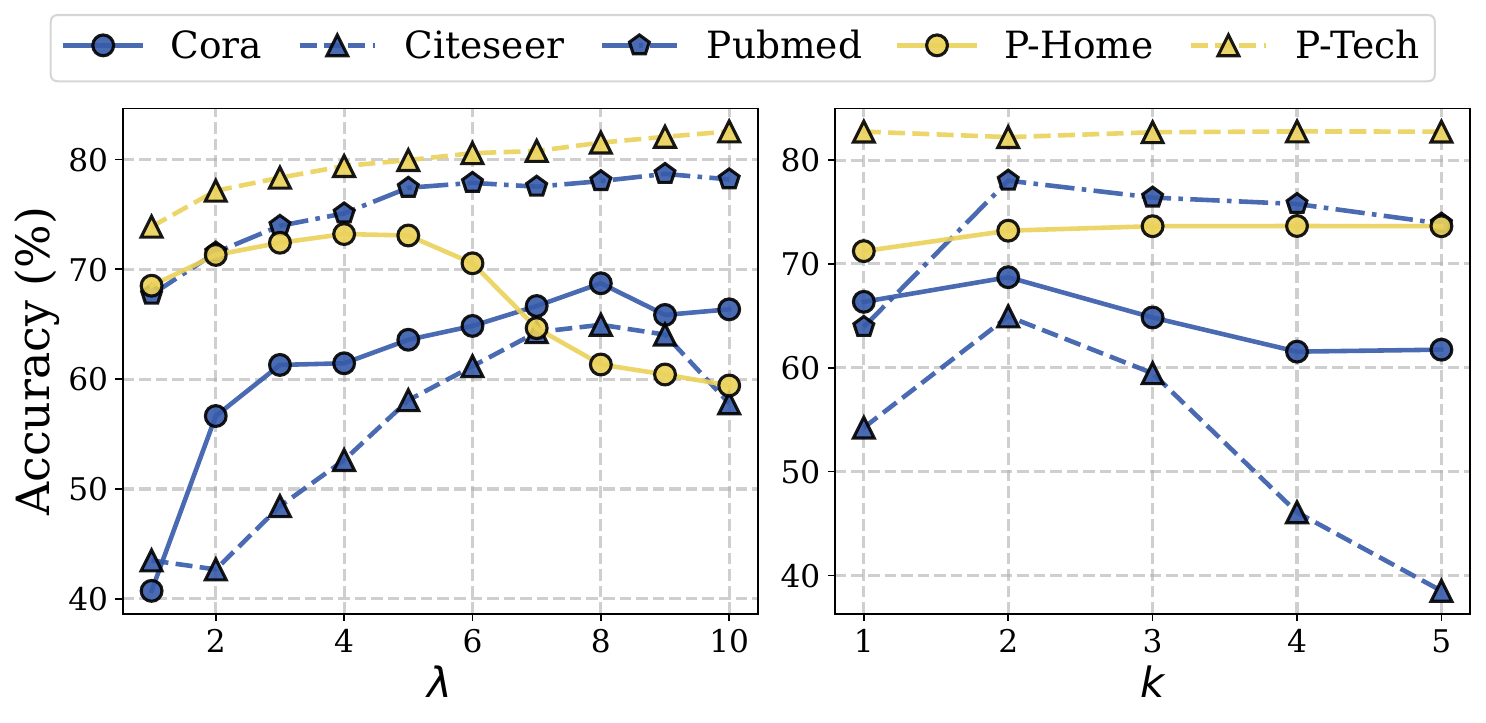}}
    \vspace{-7mm}
    \caption{Hyperparameter study of iterations $\lambda$ and the number of hops $k$.} 
    \label{fig:hyperparameter}
    \vspace{-3mm}
\end{figure}

\subsection{Hyper-parameter Sensitivity (RQ4)}

We investigate the variation of \model's performance w.r.t. $\lambda$ (the iterations of neighborhood aggregation performed) and $k$ (the number of hops for subgraph extraction during pre-training) on both citation networks and co-purchase networks, respectively. We conduct experiments with $\lambda$ varied from 1 to 10, $k$ varied from 1 to 5, and plot the results in Figure \ref{fig:hyperparameter}. Overall, our model is sensitive to both $k$ and $\lambda$. \model generally performs better with larger $\lambda$ and $k$ initially, but the performance peaks and then gradually declines. For $\lambda$, this can be attributed to an over-smoothing effect, where the features of the nodes become overly similar, making it difficult to classify. As for $k$, increasing its value leads to larger subgraphs, which can introduce overly dispersed semantics. In addition, the limitation on the number of nodes in subgraphs will result in a reduced number of subgraphs available for pre-training with larger $k$. According to Figure \ref{fig:hyperparameter}, for citation networks, the parameters lambda and k are set to 8 and 2, respectively, while for co-purchase networks, they are both set to 4.

\begin{figure}[t]
    \centering
    \resizebox{\linewidth}{!}{
    \includegraphics{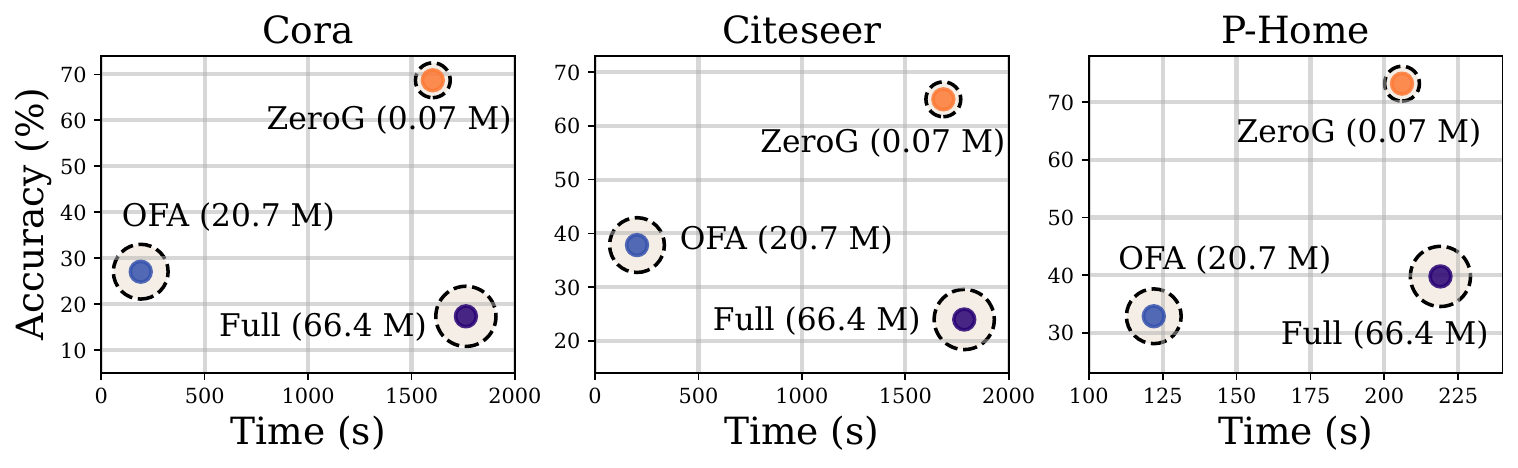}}
    \caption{Efficiency Analysis of \model.} 
    \label{fig:efficient}
    \vspace{-2mm}
\end{figure}

\subsection{Efficiency Analysis (RQ5)}

In Table \ref{fig:efficient}, we analyze the efficiency of \model, which includes the time taken to train one epoch on the Cora, Citeseer, and P-Home datasets, performance results, and the amount of parameters that require tuning for \model, OFA, and full-parameter tuning. We note that \model demands more training time than OFA \cite{liu2023one} but yields significantly better performance, presenting a trade-off between efficiency and effectiveness. In addition, \model only needs 0.07M parameters to be tuned, which is substantially less than the 20.7M required by OFA and the 66.4M needed for full-parameter tuning. Reduced parameters mitigate negative transfer in zero-shot scenarios and prevent the model from overfitting.

\subsection{Visualization}

\begin{figure}[t]
    \centering
    \resizebox{\linewidth}{!}{
    \includegraphics{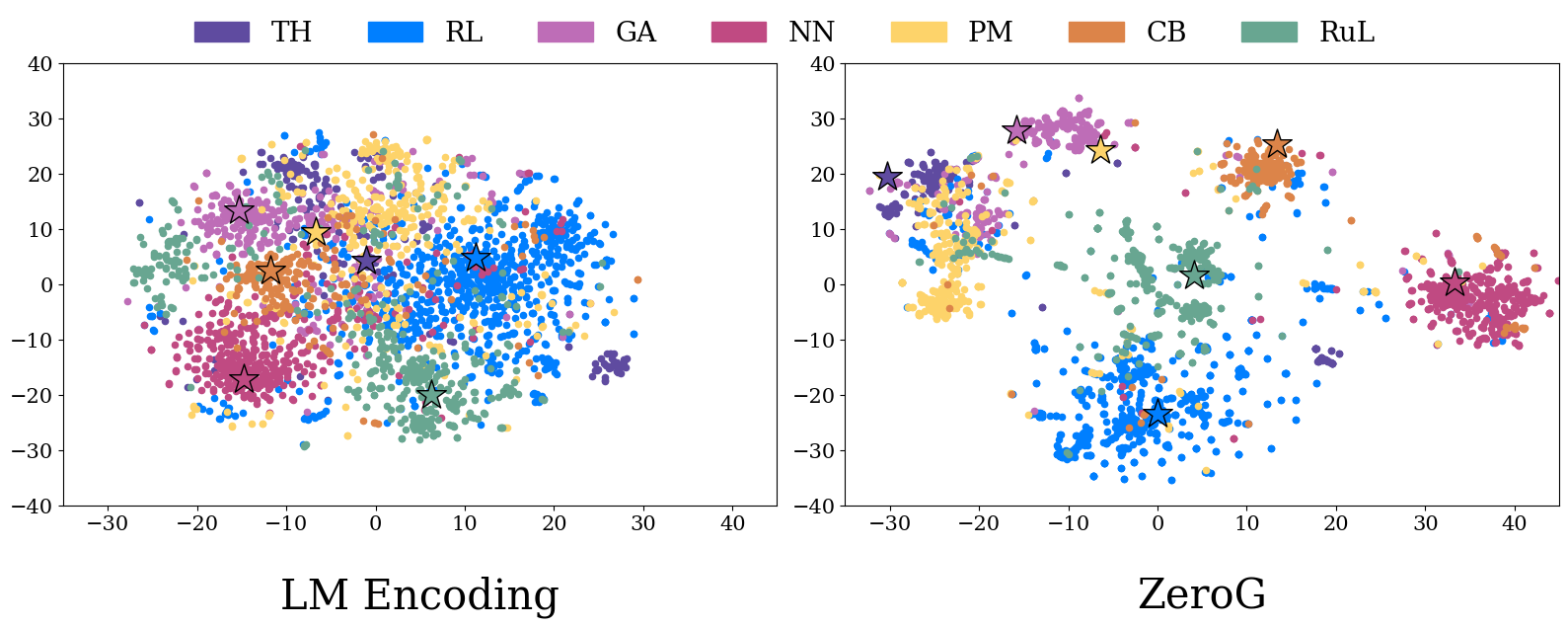}}
    \caption{Embedding visualization of Cora. Circles ($\bullet$) represent nodes, while stars ($\star$) represent classes.} 
    \label{fig:visualization}
    \vspace{-3mm}
\end{figure}

As a supplementary study of the model effectiveness, we visualize both the node and class representations with/without our proposed \model. To be specific, we use t-SNE \cite{van2008visualizing} to map the representations of Cora into two-dimensional vectors for visualization. Figure \ref{fig:visualization} shows that after equipping with our proposed \model (1) clusters of nodes with the same class (\emph{i.e.,} color in our visualization) are more cohesive in the embedding space and (2) node representations from different classes are more discriminative.
\section{Related Work}
\label{sec:related}

\subsection{Graph Prompt}

Prompt is a technique that involves augmenting foundation models with task-specific hints to help the model adopt unseen data and new tasks. It has greatly promoted the flourishing development of Artificial General Intelligence (AGI), making significant strides in the fields of natural language processing and computer vision. Different from the typical learning scheme of ``pre-training and fine-tuning'', which is based on the assumption that the pre-training stage and downstream tasks share a certain common intrinsic space, the new paradigm of ``pre-training, prompting, and fine-tuning'', which aims to reformulate input data and tasks to fit the pretext, achieves increasing attention nowadays \cite{gao2024protein}. 

This idea has also been naturally applied to the graph learning area \cite{sun2023graph}. GPPT \cite{sun2022gppt} is a pioneer work of graph prompt learning, which adopts masked edge prediction as a pre-training strategy and reformulates downstream node classification as link prediction. GraphPrompt \cite{liu2023graphprompt} unifies the downstream classification tasks and upstream link prediction pre-training into a common task template and tunes a learnable prompt token in the readout operation. In addition, ProG \cite{sun2023all} proposes a multi-task prompting framework, which unifies the format of graph prompts and language prompts. OFA \cite{liu2023one} adapts LM-embedded vectors into an NOI prompting node that contains task information to conduct various downstream tasks adaptively. However, all these studies focus on graph prompts for cross-task transfer. In this paper, we investigate the potential of applying graph prompt techniques to cross-dataset transfer, utilizing a hard prompting node to provide unique identifiers and additional general semantics for both source and target datasets.

\subsection{LM for Graphs}

Recently, language models (LMs) have been increasingly utilized in graph-related tasks \cite{shen2021entity,shu2022tiara,li2022community}. Current methodologies can be classified into three categories, distinguished by the role of LMs in graph-related tasks: as the enhancer, predictor, and alignment component \cite{li2023survey}. Specifically, LM-as-enhancer approaches correspond to enhancing the quality of node embeddings with the help of powerful LMs. For example, TAPE \cite{he2023explanations} utilizes LLMs to enrich initial node embeddings in GNNs with semantic knowledge relevant to the nodes, notably improving embedding quality. WalkLM \cite{tan2023walklm} generates roughly meaningful textual sequences by an automated textualization program and fine-tunes an LM using them. In addition, taking LM as predictors, several works \cite{wang2023can,zhao2023graphtext,ye2023natural,fatemi2023talk,chen2024graphwiz} utilize LMs to directly make predictions for a wide range of graph tasks, such as classifications and reasonings, within a unified generative paradigm. 
Lastly, GNN-LM alignment ensures that each encoder's unique functionalities are preserved while coordinating their embedding spaces at a specific stage. For example, MoMu \cite{su2022molecular}, MoleculeSTM \cite{liu2023multi}, ConGraT \cite{brannon2023congrat}, and RLMRec \cite{wei2023llmrec} merge GNN and LM embeddings, enriching graphs with textual knowledge to boost reasoning.
In \modelv, we follow the approach of using LM as an enhancer, employing the LM for unified graph representation learning to bridge the semantic gap between feature and label spaces across source and target graphs.

\section{Conclusion and Future Work}
\label{sec:conclusion}

In this paper, we systematically summarize and analyze cross-dataset zero-shot transfer for graph data and introduce \textsc{ZeroG}, a novel approach to address this task. The key design of our model involves unifying graph representations via a language model, using the prompt-based subgraph sampling strategy to embed graph structures and semantics into pre-training datasets, and employing a lightweight fine-tuning approach to minimize the risk of overfitting. Comprehensive experiments validate the effectiveness of our model. \textsc{ZeroG} reveals great potential as the future foundation model on the graph. Future work will explore expanding \textsc{ZeroG}'s capabilities to include link-level and graph-level tasks and incorporate regression tasks to broaden the applicability.

\begin{acks}
This work was supported by NSFC Grant No. 62206067, HKUST-HKUST(GZ) 20 for 20 Cross-campus Collaborative Research Scheme C019 and Guangzhou-HKUST(GZ) Joint Funding Scheme 2023A03J0673.
\end{acks}




\bibliographystyle{ACM-Reference-Format}
\bibliography{mybib.bib}

\appendix
\begin{table*}[t]
\caption{Example of node features and class descriptions}
\label{tab:des}
\begin{tabular}{l|l}
\toprule
Dataset & Node feature \\ \midrule \midrule
\multirow{5}{*}{Cora} & Learning sparse perceptrons. We introduce a new algorithm designed to learn sparse perceptrons over input representations  \\
 & which include high-order features. Our algorithm, which is based on a hypothesis-boosting method, is able to PAC-learn a\\
 & relatively natural class of target concepts. Moreover, the algorithm appears to work well in practice: on a set of three problems \\
 & domains, the algorithm produces classifiers that utilize small numbers of features yet exhibit good generalization performance.\\
 &  Perhaps most importantly, our algorithm generates concept descriptions that are easy for humans to understand.  \\ \midrule \midrule
Dataset & Class description \\ \midrule  
\multirow{7}{*}{Cora} & Theory. The ``Theory'' category likely refers to research papers that delve into the theoretical aspects of machine learning \\
 & and artificial intelligence. This includes a broad array of topics such as theoretical foundations of various machine learning \\
 & algorithms, performance analysis, studies on learning theory, statistical learning, information theory, and optimization \\
 & methods. Additionally, it could encompass the development of new theoretical frameworks, investigations into the essence  \\
 &of intelligence, the potential for artificial general intelligence, as well as the ethical implications surrounding AI. Essentially,  \\
 & the ``Theory'' category encapsulates papers that primarily focus on theoretical concepts and discussions, contrasting with  \\
 & more application-oriented research which centers on specific techniques and their practical implementation. \\ \bottomrule
\end{tabular}
\end{table*}

\section{Dataset Construction}
\label{app:datasets}
We constructed two datasets, P-Home and P-Tech, both of which were extracted from the ogbn-products dataset. We use raw texts processed by Chen et al. \cite{chen2023exploring} as the description of nodes and labels. The ogbn-products dataset serves as a large-scale, undirected, and unweighted graph that encapsulates an Amazon product co-purchasing network. Within this graph, nodes are products available on Amazon, while the edges signify co-purchasing relationships, indicating that two products have been bought together by customers. The features of the nodes are the descriptions of the products. Since each product (node) belongs to only one category, our method of extracting based on category ensures that there is no overlap between P-Home and P-Tech.

\textbf{P-Home:}
For P-Home, we selected categories related to household items: ``Baby Products'', ``Appliances'', ``All Beauty'', ``Office \& School Supplies'', and ``Home Improvement''. The number of nodes in each category is as follows: 3653, 3024, 1969, 630, and 514.

\textbf{P-Tech:} For P-Tech, we chose products related to technology, including ``Software'', ``Video Games'', and ``Industrial \& Scientific''. The number of nodes of each category is 3079, 20911, and 17438.

\section{Texual Information}
\label{app:description}

\subsection{Node Features and Class Descriptions}
\label{app:node_des}
We follow OFA \cite{liu2023one} to collect the node features and class descriptions for all datasets. Taking citation networks as examples, the node features are the titles and abstracts of papers. The class descriptions of citation networks are the class names with detailed descriptions. An example from Cora is illustrated in Table \ref{tab:des}.

\subsection{Dataset Descriptions}
\label{app:dataset_des}
In this section, we will enumerate the description of each dataset utilized, which is also used for prompt texts. Generally, we describe the classes and average degree of each dataset. The dataset descriptions are generated by GPT-4 and subsequently manually refined.
\subsubsection{Cora}
The Cora \cite{yang2016revisiting} dataset is a fundamental resource in the field of graph learning, particularly within the realm of machine learning research. It represents a network of scientific publications. There are 7 categories in Cora: Theory, covering theoretical aspects of machine learning and AI; Reinforcement Learning, including research on reinforcement learning, a type of machine learning; Genetic Algorithms, dealing with genetic algorithms, a type of optimization algorithm inspired by natural evolution. Neural Networks, focusing on artificial neural networks, a subset of machine learning. Probabilistic Methods, pertaining to research on probabilistic methods in machine learning, using probability mathematics to handle uncertainty and make predictions. Case Based, focusing on case-based reasoning in AI, a method that solves new problems by referring to similar past cases. Rule Learning, involving the generation of rules for decision-making systems. The average degree of Cora is 4.

\subsubsection{Citeseer}
The Citeseer \cite{yang2016revisiting} dataset is a prominent academic resource in the field of computer science, categorizing publications into six distinct areas. These are Agents, focusing on intelligent agents; Machine Learning (ML), covering all aspects of learning techniques and applications; Information Retrieval (IR), dealing with data and text indexing and retrieval; Databases (DB), related to database management and data mining; Human-Computer Interaction (HCI), emphasizing computer technology interfaces for humans; and Artificial Intelligence (AI), a broad category encompassing general AI theory and applications, excluding certain subfields. The average degree of this graph is 2.
\subsubsection{Pubmed}
The PubMed \cite{yang2016revisiting} dataset comprises three categories: Experimental studies on diabetes mechanisms and therapies, Type 1 Diabetes research focusing on autoimmune processes and treatments, and Type 2 Diabetes studies emphasizing insulin resistance and management strategies. Each category addresses specific aspects of diabetes research, aiding in understanding and treating this complex disease. The average degree of this graph is 4.5.
\subsubsection{Arxiv}
The arXiv dataset is a notable resource in the field of graph learning, particularly in the area of computer science research. This dataset forms a directed graph representing the citation network among all Computer Science papers on arXiv, as indexed by the Microsoft Academic Graph (MAG). Each node in this network corresponds to a paper, and directed edges indicate citations. The dataset's primary challenge is predicting the 40 subject areas of arXiv CS papers, such as cs.AI, cs.LG, and cs.OS. The task is structured as a 40-class classification problem.

\subsubsection{P-Home}
This graph is of amazon products about home using. There are six categories. Baby Products: A category dedicated to items designed for infants and toddlers, including hygiene, feeding, and skin care essentials; Appliances: This section features electrical machines and devices intended for household tasks, such as cooking, cleaning, and food preservation; All Beauty: A broad range of personal care products aimed at enhancing or maintaining physical appearance and hygiene; Office \& School Supplies: Items and tools used for writing, organizing, and conducting daily activities in educational and professional settings; Home Improvement: Products and materials focused on repairing, enhancing, or maintaining the functionality and aesthetics of living spaces. The average degree of this graph is 26.93.

\subsubsection{P-Tech}
This graph is of amazon products about technologies. There are three categories. Software: Computer programs and applications developed to perform specific tasks on computing devices, ranging from productivity to creative design; Video Games: Interactive entertainment software and accessories designed for recreational play on consoles, computers, and portable devices; Industrial \& Scientific: Equipment, tools, and materials used in industrial operations and scientific research, including measurement, fabrication, and experimental applications. The average degree of this graph is 87.60.

\subsubsection{Wiki-CS}
The Wiki-CS \cite{mernyei2020wiki} dataset is a comprehensive collection of Wikipedia entries, systematically categorized into ten distinct areas of computer science. These categories include Computational Linguistics, focusing on the intersection of computer science and linguistics; Databases, covering database technologies and theories; Operating Systems, detailing the software that manages computer hardware; Computer Architecture, exploring the design and structure of computer systems; Computer Security, addressing the protection of information systems; Internet Protocols, discussing the rules governing internet data exchange; Computer File Systems, about methods for storing and organizing computer files; Distributed Computing Architecture, concerning computations spread across multiple machines; Web Technology, focusing on the technologies underpinning the web; and Programming Language Topics, which includes various aspects of programming languages. This dataset serves as a valuable resource for understanding diverse computer science topics as represented in Wikipedia, reflecting the breadth and depth of the field.

\end{document}